\documentclass[10pt,twocolumn,letterpaper]{article}
\usepackage{cvpr}
\usepackage[T1]{fontenc}
\usepackage{multirow}
\usepackage{makecell}
\usepackage{colortbl}
\definecolor{rankfirst}{HTML}{FADADD}
\definecolor{ranksecond}{HTML}{DCEEFF}
\definecolor{rankthird}{HTML}{E2F0D9}
\newlength{\rankcolumnwidth}
\newcommand{\rankcell}[2]{%
  \makebox[\dimexpr\rankcolumnwidth-2\tabcolsep\relax][c]{%
    {\setlength{\fboxsep}{0pt}%
     \colorbox{#1}{\makebox[0.8\rankcolumnwidth][c]{\strut\textbf{#2}}}}}}
\definecolor{cvprblue}{rgb}{0.21,0.49,0.74}
\usepackage[pagebackref,breaklinks,colorlinks,allcolors=cvprblue,hypertexnames=false]{hyperref}
\def\confName{CVPR}
\def\confYear{2027}
\title{DAUPNet: Domain-Aware Uncertainty Modeling for Reliable Prototype Discrimination in Cross-Domain Few-Shot Semantic Segmentation}
\author{
Lei Yuan$^{1}$ \quad
Zhongxu Hu$^{1}$ \quad
Jingyi Wen$^{1}$ \quad
Pengxing Yi$^{1,*}$\\
$^{1}$Department of Mechanical Science and Engineering,\\
Huazhong University of Science and Technology, China\\
{\ttfamily\small madness\_lei@163.com, zhongxu\_hu@hust.edu.cn}\\
{\ttfamily\small jingyi\_wen2003@163.com, pxyi@hust.edu.cn}\\
{\small $^{*}$Corresponding author}
}
\begin{document}
\maketitle
\begin{abstract}
Cross-domain few-shot semantic segmentation (CD-FSS) has predominantly been formulated as learning domain-invariant representations or improving support-query correspondence. Nevertheless, large domain shifts still make prototype matching unreliable: inconsistent hierarchical responses corrupt the support representation, deterministic prototypes cannot express boundary and appearance ambiguity, and treating prototypes with different reliability equally during optimization weakens foreground-background separation. DAUPNet is therefore proposed as a unified framework that reformulates cross-domain prototype matching as uncertainty-aware prototype discrimination. DAUPNet first harmonizes hierarchical support-query features to provide stable evidence, then represents foreground and background prototypes probabilistically, and finally uses their estimated uncertainty to regulate contrastive optimization. On four standard target domains, DAUPNet achieves 73.39\% and 77.01\% average mIoU in the 1-shot and 5-shot settings, respectively, including substantial gains on the two medical domains. These results support combining multi-level prototype statistics with dispersion-weighted optimization for CD-FSS under severe domain shift. 
Code will be released upon acceptance.
\end{abstract}

\section{Introduction}
\begin{figure}[t]
\centering
\scalebox{1}[0.8]{\includegraphics[width=\linewidth]{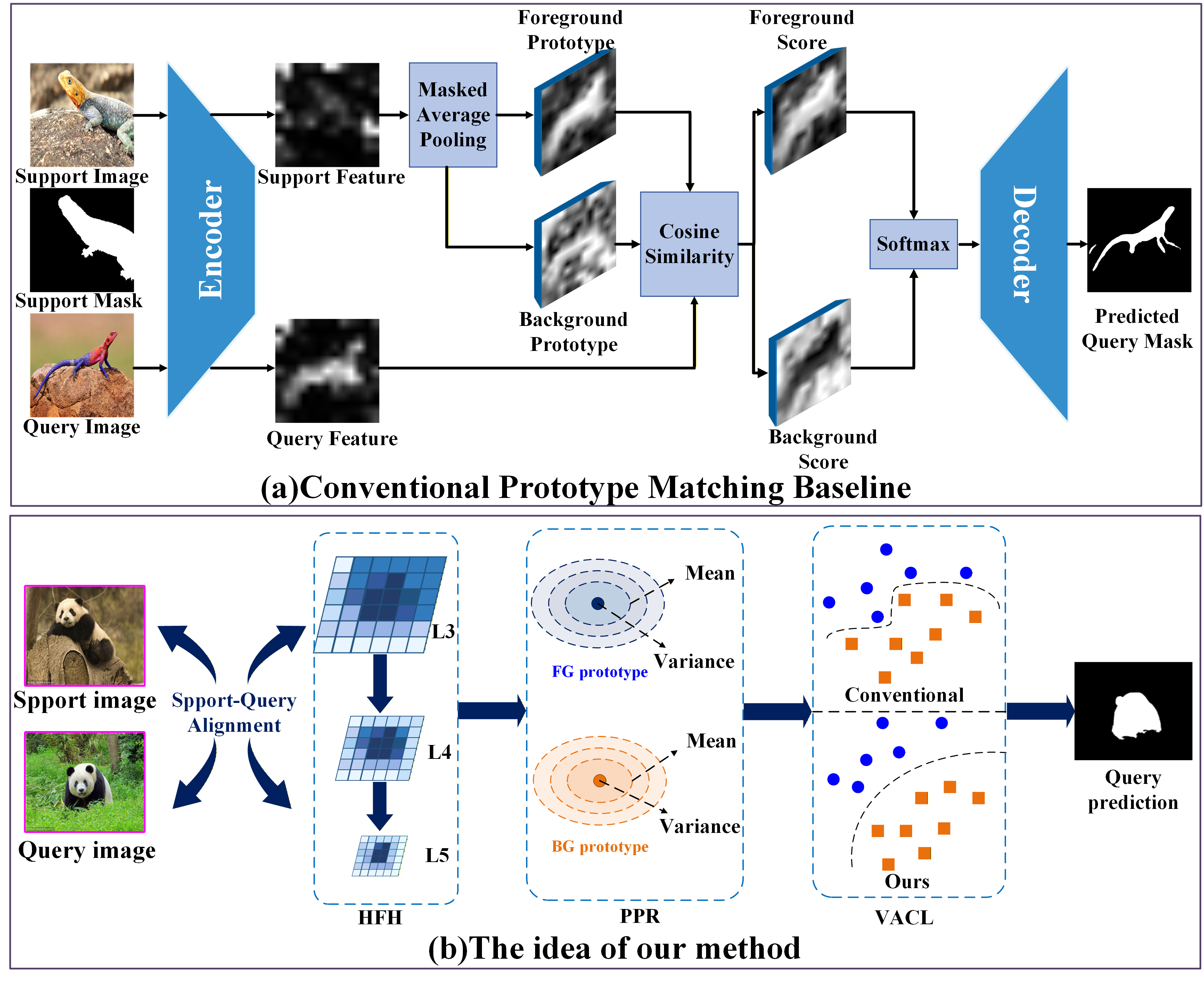}} 
\caption{Conceptual overview. (a) Deterministic foreground/background prototype matching. (b) HFH, PPR, and VACL improve prototype discrimination; VACL is used only during training.}
\label{fig:fig1}  
\end{figure}
Few-shot semantic segmentation (FSS) segments novel classes from only a few annotated support images, reducing the dependence of conventional segmentation models on dense annotations and fixed training categories. In practical applications, however, support and query images may come from domains that differ markedly from the source training data in acquisition modality, texture, contrast, and object structure. Cross-domain few-shot semantic segmentation (CD-FSS) addresses this setting, but the combined scarcity of annotations and domain shift makes reliable support-query transfer particularly difficult. Existing studies have approached this difficulty through domain-invariant representation \cite{PATNet}, target-domain adaptation and support-query correspondence mining \cite{IFA,DR-Adapter} , and multi-scale or low-level feature enhancement \cite{APM,LoEC,HSL}. Despite this progress, segmentation remains fragile when domain shift undermines the reliability of the prototypes on which support-query matching depends.

This prototype-reliability problem extends beyond feature alignment and arises throughout prototype construction and discrimination. First, separately processing high-level semantics and low-level appearance leaves cross-level evidence inconsistent, so the representation pooled into a prototype may already be unstable. Second, masked average pooling compresses that evidence into a deterministic vector and therefore cannot indicate whether the support region has an ambiguous boundary, weak contrast, or heterogeneous appearance. Third, conventional contrastive optimization treats prototype combinations with different uncertainty as equally reliable, allowing unstable comparisons to disproportionately shape the foreground-background decision boundary. These effects are coupled: misaligned features produce uncertain prototypes, and optimizing them without accounting for their reliability further amplifies unstable discrimination.

Fig.\ref{fig:fig1} contrasts conventional prototype matching with the uncertainty-aware formulation of DAUPNet. The conventional path pools masked support features into deterministic foreground/background prototypes for query matching. DAUPNet instead preserves a reliability chain: hierarchical harmonization stabilizes the evidence used to form prototypes; probabilistic representation exposes residual foreground and background uncertainty; and variance-aware optimization weights prototype comparisons using dispersion-derived scores.

Based on this view, DAUPNet is proposed as a unified prototype learning framework organized around reliable uncertainty-aware discrimination rather than a collection of independent enhancements. DAUPNet retains the simplicity of prototype-based segmentation and implements the reliability chain in three successive stages. Hierarchical Feature Harmonization (HFH) coordinates multi-level support-query features so that prototype estimation starts from consistent cross-domain evidence. Probabilistic Prototype Representation (PPR) then represents foreground and background by prototype means and dispersion-derived scores, retaining information discarded by point prototypes. Finally, Variance-Aware Contrastive Learning (VACL) converts the variance-derived uncertainty into reliability weights, weighting prototype comparisons according to their dispersion scores to encourage foreground-background separation. Thus, the components connect feature construction, prototype statistics, and training supervision toward reliable prototype discrimination under domain shift. 

The main contributions are:
\begin{itemize}
\item{This study identifies unreliable prototype discrimination as a central bottleneck in CD-FSS and formulates its coupled causes across feature construction, uncertainty representation, and decision-margin optimization. This formulation provides a coherent alternative to treating domain shift solely as feature alignment.
}
\item{DAUPNet establishes an end-to-end reliability chain from hierarchical feature harmonization to probabilistic foreground-background prototypes and variance-aware contrastive optimization. HFH stabilizes the evidence entering prototype estimation, while PPR and VACL expose and control the remaining uncertainty during representation and learning.}
\item {The results show that uncertainty-aware prototype discrimination is especially valuable under severe appearance and modality shifts. DAUPNet improves the baseline by 7.66 mIoU points on average in the 1-shot setting and reaches 73.39\%/77.01\% average mIoU across four target domains, with particularly strong performance on weak-boundary and grayscale medical images.}
\end{itemize}

\begin{figure*}[t]
\centering
\includegraphics[width=\linewidth]{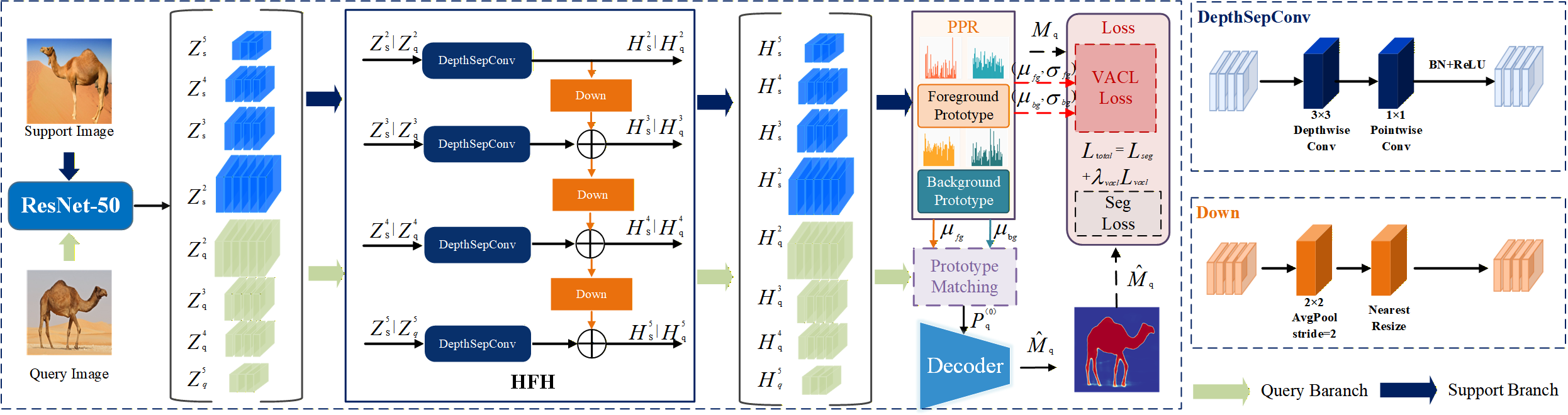} 
\caption{Architecture of DAUPNet.}
\label{fig:fig2DAUPNET}  
\end{figure*}

\section{Related Work}
\subsection{Few-Shot Semantic Segmentation}
FSS is primarily built upon prototype matching, dense correspondence modeling, and relational reasoning. PANet\cite{PANet} establishes the foundational paradigm of masked average pooling and cosine matching. SSP\cite{SSP} enhances representational capacity via query self-support prototypes. Subsequent work has improved discriminability through multi-prototype representation, cross-scale aggregation, and boundary enhancement. However, these deterministic prototypes assume stable feature distributions and clear boundaries---assumptions that break under domain shift. Overall, same-domain FSS has matured considerably. Yet most methods implicitly assume similar training-testing distributions. When significant modality discrepancies exist between source and target domains, their matching mechanisms often fail to maintain stable generalization performance \cite{FSS}.

\subsection{Cross-Domain Few-Shot Semantic Segmentation}

CD-FSS simultaneously faces category scarcity and domain shift, making it a substantially harder yet more realistic extension.

Domain-Invariant Feature Learning. PATNet\cite{PATNet} pioneered the CD-FSS benchmark and proposed domain-invariant feature transformation. DMTN introduced dual matching for high-level semantics and low-level textures \cite{DMTN}. However, these methods focus on feature alignment while neglecting prototype instability under domain shift.

Target-Domain Adaptation. IFA\cite{IFA} improved adaptability by iteratively mining support-query correspondences. ABCDFSS\cite{ABCDFSS} advocated test-time adaptation before comparison. Yet these iterative strategies still operate on deterministic prototypes that cannot capture cross-domain ambiguity.

Multi-Scale Enhancement. DR-Adapter\cite{DR-Adapter}, APM\cite{APM}, and LoEC\cite{LoEC} improve robustness via domain perturbation, frequency suppression, and low-level calibration. HSL\cite{HSL} explores hierarchical semantic learning.

HSL\cite{HSL} uses prototype confidence to adapt inference thresholds, while LoEC\cite{LoEC} calibrates predictions using low-level information. In DAUPNet, dispersion scores derived from multi-level foreground/background support statistics weight prototype contrastive relations during training, while query matching uses the prototype means. This design connects hierarchical feature aggregation, prototype statistics, and contrastive optimization within a shared pipeline.

\section{Method}
\subsection{Problem Definition and Notation}
This paper investigates CD-FSS. Let the source-domain training set be denoted as $\mathcal{D}_{src}$ and the target-domain test set as $\mathcal{D}_{tgt}$. The two domains have different image distributions, i.e., $P_{src}(I)\neq P_{tgt}(I)$, and their training and testing categories are mutually disjoint. Each episode contains a support set and a query image. Under the 1-way $K$-shot setting, the support set is defined as
\begin{equation}
\mathcal{S}=\{(I_s^k,M_s^k)\}_{k=1}^{K}
\label{eq:body-1}
\end{equation}

where $I_s^k$ and $M_s^k$ denote the $k$-th support image and its binary mask, respectively. The query image is denoted as $I_q$. During training, the ground-truth query mask $M_q$ is available for computing the segmentation loss, whereas during inference only $I_q$ is given. The goal is to predict the query mask $\hat{M}_q$ conditioned on $\mathcal{S}$.

Let $b\in\{s,q\}$ denote the support or query branch, and let $l\in\mathcal{L}$ denote the feature level. The implementation uses $\mathcal{L}=\{2,3,4,5\}$. The deepest layer is denoted as $l_d=\max(\mathcal{L})$. After channel unification, each feature level has dimension $d=256$, and the feature at level $l$ is written as
\begin{equation}
X_b^l\in\mathbb{R}^{B\times d\times h_l\times w_l}\label{eq:body-2}
\end{equation}

where $B$ denotes the batch size and $h_l\times w_l$ denotes the spatial size.

\subsection{Overall Framework} 
As shown in Fig.\ref{fig:fig2DAUPNET}, DAUPNet realizes uncertainty-aware prototype discrimination as a sequential reliability chain comprising HFH, PPR, and VACL. The support image $I _ { s }$ and query image $I _ { q }$ are first fed into a shared encoder to extract multi-level features:
\begin{equation}
\{F_s^l\}_{l\in\mathcal{L}}=E(I_s),\qquad
\{F_q^l\}_{l\in\mathcal{L}}=E(I_q)\label{eq:body-3}
\end{equation}

A Channel Reduce module projects features from different backbone stages into a unified feature space:
\begin{equation}
Z_b^l=\operatorname{CR}^l(F_b^l),\qquad b\in\{s,q\},\ l\in\mathcal{L}\label{eq:body-4}
\end{equation}

HFH then coordinates the support and query features across hierarchical levels and outputs harmonized features:
\begin{equation}
\{H_s^l\}_{l\in\mathcal{L}},\quad \{H_q^l\}_{l\in\mathcal{L}}\label{eq:body-5}
\end{equation}

Based on the support harmonized features and the support mask, PPR estimates foreground and background probabilistic prototypes:
\begin{equation}
(\mu_{fg},\sigma_{fg}),\quad(\mu_{bg},\sigma_{bg})\label{eq:body-6}
\end{equation}

Here, $\mu$ denotes the prototype mean, while $\sigma$ is a variance-derived estimate of prototype uncertainty. The foreground/background prototype means are matched with the deepest query feature $H _ { q } ^ { l _ { d } }$ to obtain a low-resolution prediction $P _ { q } ^ { ( 0 ) }$ . The decoder then fuses this prediction with multi-level query features and progressively upsamples the result:
\begin{equation}
\hat{M}_q=\operatorname{Decoder}(P_q^{(0)},\{H_q^l\}_{l\in\mathcal{L}})\label{eq:body-7}
\end{equation}

During training, the segmentation loss is computed using the query mask $M _ { q }$, and VACL uses the variance-derived uncertainty estimated by PPR to impose a reliability-weighted contrastive constraint on batch-level probabilistic prototypes. During inference, only prototype matching and decoding are performed, so the contrastive optimization introduces no additional inference branch.

\subsection{Hierarchical Feature Harmonization Module}
HFH is the first stage of the reliability chain. It reduces shallow texture noise and integrates local detail with deeper semantic information before prototype statistics are computed. Given the channel-reduced features $Z _ { s } ^ { l }$ and $Z _ { q } ^ { l }$, HFH outputs harmonized features $H _ { s } ^ { l }$ and $H _ { q } ^ { l }$. Rather than directly performing segmentation, HFH stabilizes the multi-level evidence from which PPR estimates prototype means and uncertainty.

As shown in Fig.\ref{fig:fig2DAUPNET}, HFH follows a hierarchical coordination process consisting of intra-level processing and bottom-up additive fusion. Shallow levels focus on local texture and boundary information, while deeper levels emphasize structural semantics. At each level, a depthwise separable convolution refines the reduced feature:
\begin{equation}
\widetilde{Z}_b^l=\operatorname{DepthSepConv}_l(Z_b^l),\qquad b\in\{s,q\}.
\label{eq:hdh_intra}
\end{equation}
Here, $\operatorname{DepthSepConv}_l$ comprises a $3\times3$ depthwise convolution, a $1\times1$ pointwise convolution, batch normalization, and ReLU. The support and query branches use the same processing parameters.

The bottom-up path transfers boundary and texture information from shallower levels to deeper levels. For the current level $l$, the preceding shallower output is downsampled and spatially aligned to the current feature size:
\begin{equation}
D_b^{l_p}=\operatorname{Down}_{l_p\rightarrow l}(H_b^{l_p}),
\label{eq:hdh_down}
\end{equation}
where $l_p$ denotes the preceding shallower level and $\operatorname{Down}$ applies $2\times2$ average pooling with stride 2, followed by nearest-neighbor resizing. The harmonized feature is then computed by additive fusion:
\begin{equation}
H_b^l=\begin{cases}
\widetilde{Z}_b^l, & l=\min(\mathcal{L}),\\
\widetilde{Z}_b^l+D_b^{l_p}, & l>\min(\mathcal{L}).
\end{cases}
\label{eq:hdh_fusion}
\end{equation}

Through these operations, HFH combines boundary information and semantic context within a shared support-query feature hierarchy. The support features ${H_s^l}$ are used for probabilistic prototype estimation, the deepest query feature $H_q^{l_d}$ is used for prototype matching, and the multi-level query features ${H_q^l}$ are used by the decoder for spatial recovery.

\subsection{Probabilistic Prototype Representation Module}
\begin{figure}[t]
\centering
\includegraphics[width=\linewidth]{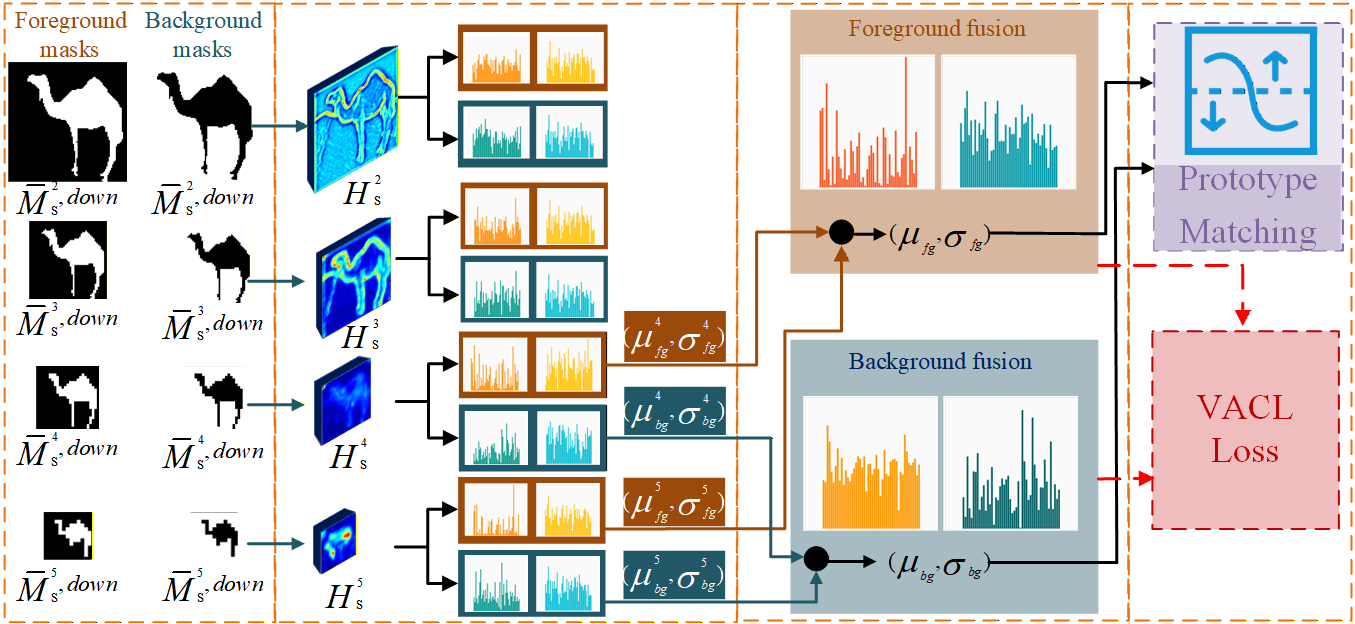}
\caption{PPR fuses foreground/background means and variance-derived scores from the two deepest support levels. Matching uses only the means; VACL also uses the scores during training. Shallow levels show the feature hierarchy.}
\label{fig:ppr_structure}
\end{figure}
As shown in Fig.\ref{fig:ppr_structure}, PPR is the second stage of the reliability chain. It converts the harmonized support evidence into probabilistic foreground/background prototypes through mask resizing, level-wise statistics, and fusion of the two deepest features. Each prototype retains a class center and a nonnegative dispersion score, which serves as an uncertainty proxy for contrastive weighting. Given the support HFH features $\{H_s^l\}_{l\in\mathcal{L}}$ and the support mask $M _ { s }$, the mask is first resized to each of the two selected feature levels, yielding $M _ { s } ^ { l }$. The background mask is defined as
\begin{equation}
M_{bg}^l=1-M_s^l\label{eq:body-8}
\end{equation}

The foreground and background pixel numbers are
\begin{equation}
n_{fg}^l=\sum_x M_s^l(x),\qquad
n_{bg}^l=\sum_x M_{bg}^l(x)\label{eq:body-9}
\end{equation}

where both values are clamped in the implementation to avoid division by zero. The foreground and background prototype means are computed as
\begin{equation}
\mu_{fg}^l=\frac{\sum_x H_s^l(x)M_s^l(x)}{n_{fg}^l},
\qquad
\mu_{bg}^l=\frac{\sum_x H_s^l(x)M_{bg}^l(x)}{n_{bg}^l}\label{eq:body-10}
\end{equation}

To quantify the uncertainty concealed by a point prototype, PPR computes foreground and background feature variances:
\begin{equation}
\begin{aligned}
v_{fg}^l=
\frac{\sum_x (H_s^l(x)-\mu_{fg}^l)^2M_s^l(x)}{n_{fg}^l},
\\
v_{bg}^l=
\frac{\sum_x (H_s^l(x)-\mu_{bg}^l)^2M_{bg}^l(x)}{n_{bg}^l}
\label{eq:body-11}
\end{aligned}
\end{equation}

A learned correction layer maps these statistics to nonnegative scores, clipped at 50. Here, probabilistic representation refers to the mean--dispersion pair; the score is not a calibrated posterior variance:
\begin{equation}
\begin{aligned}
\sigma_{fg}^l=\min\{\operatorname{Softplus}(\operatorname{Linear}(v_{fg}^l)),50\},
\\
\sigma_{bg}^l=\min\{\operatorname{Softplus}(\operatorname{Linear}(v_{bg}^l)),50\}
\end{aligned}
\label{eq:body-12}
\end{equation}

When multiple feature levels are available, PPR fuses the deepest level $l_d$ and second-deepest level $l_{sd}$ to balance semantic abstraction and spatial detail:
\begin{equation}
\mu_{fg}=0.4\mu_{fg}^{l_{sd}}+0.6\mu_{fg}^{l_d},
\qquad
\mu_{bg}=0.4\mu_{bg}^{l_{sd}}+0.6\mu_{bg}^{l_d}\label{eq:body-13}
\end{equation}

The correction layer is optimized jointly through VACL. The dispersion scores are aggregated using a harmonic mean:
\begin{equation}
\begin{aligned}
\sigma_{fg}=
\frac{2}{\frac{1}{\sigma_{fg}^{l_{sd}}+\varepsilon}+
\frac{1}{\sigma_{fg}^{l_d}+\varepsilon}},
\\
\sigma_{bg}=
\frac{2}{\frac{1}{\sigma_{bg}^{l_{sd}}+\varepsilon}+
\frac{1}{\sigma_{bg}^{l_d}+\varepsilon}}
\label{eq:body-14}
\end{aligned}
\end{equation}

Thus, PPR produces the foreground prototype $( \mu _ { f g } , \sigma _ { f g } )$ and background prototype $( \mu _ { b g } , \sigma _ { b g } )$. The means are used for query matching, while both the means and their variance-derived uncertainty estimates are passed to VACL, connecting probabilistic representation directly to reliability-aware optimization. In the K-shot setting, prototypes are first extracted from each support sample and then averaged to obtain episode-level probabilistic prototypes.

Given the deepest query feature $H_q^{l_d}$, prototype matching normalizes the query feature and prototype means:
\begin{equation}
\begin{aligned}
\bar{H}_q^{l_d}&=\operatorname{Norm}(H_q^{l_d}),\\
\bar{\mu}_{fg}&=\operatorname{Norm}(\mu_{fg}),\quad
\bar{\mu}_{bg}=\operatorname{Norm}(\mu_{bg})
\label{eq:body-15}
\end{aligned}
\end{equation}

The foreground and background similarities are computed as
\begin{equation}
s_{fg}(x)=\bar{H}_q^{l_d}(x)^\top \bar{\mu}_{fg},
s_{bg}(x)=\bar{H}_q^{l_d}(x)^\top \bar{\mu}_{bg}
\label{eq:body-16}
\end{equation}

The low-resolution foreground probability is then obtained by softmax:
\begin{equation}
P_q^{(0)}(x)=
\operatorname{Softmax}([s_{fg}(x),s_{bg}(x)])_{fg}\label{eq:body-17}
\end{equation}

The decoder fuses $P_q^{(0)}$ with the multi-level query HFH features and upsamples the result to the original image size:
\begin{equation}
\hat{M}_q=\operatorname{Sigmoid}(\operatorname{Conv}_{1\times1}(\operatorname{Up}(X^{l_{min}})))\label{eq:body-18}
\end{equation}

Here, $X^{l_{min}}$ is the fused query feature at the shallowest decoder level. This design preserves the discriminative semantics of prototype matching while using shallow query features to recover spatial boundaries.

\subsection{Variance-Aware Contrastive Learning Module}
\begin{figure}[htbp]
\centering
\includegraphics[width=\linewidth]{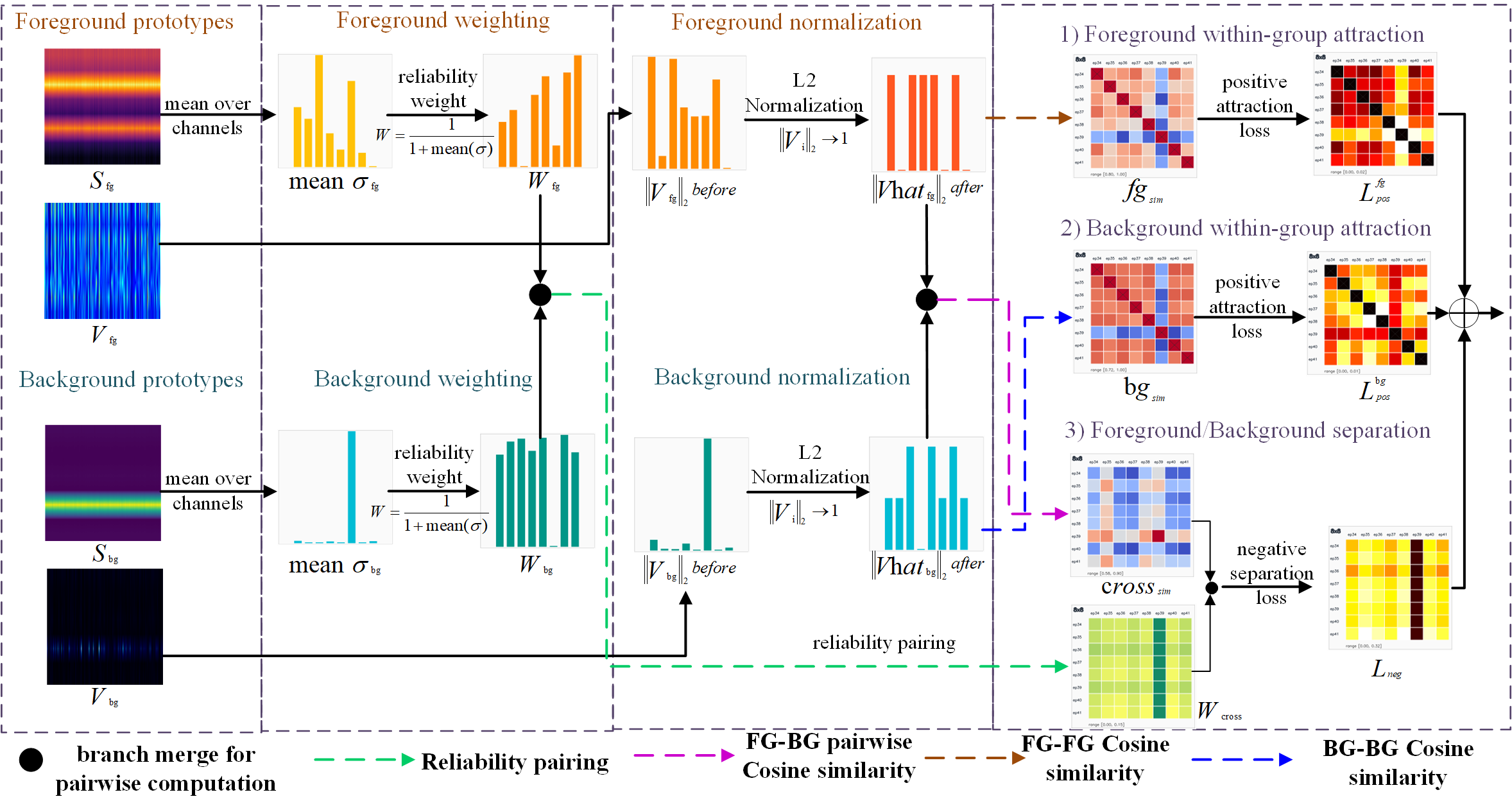} 
\caption{Variance-Aware Contrastive Learning (VACL) for reliability-guided prototype discrimination.}
\label{VACL}  
\end{figure}
As shown in Fig.\ref{VACL}, VACL completes the reliability chain by converting the variance-derived uncertainty from PPR into optimization weights. It collects batch-level foreground/background probabilistic prototypes, normalizes their representations, and constructs within-group attraction and foreground-background separation losses. Prototypes with lower dispersion scores receive larger weights in within-group attraction and foreground-background separation, while comparisons involving larger scores receive lower weights. Within a batch, foreground and background prototype means are concatenated as
\begin{equation}
\begin{aligned}
V_{fg}=\operatorname{Concat}(\mu_{fg}^{(1)},\ldots,\mu_{fg}^{(N)}),\\
V_{bg}=\operatorname{Concat}(\mu_{bg}^{(1)},\ldots,\mu_{bg}^{(N)})
\end{aligned}
\label{eq:body-19}
\end{equation}

where $N$ is the number of episode-level prototype pairs in the batch, with corresponding dispersion scores
\begin{equation}
\begin{aligned}
S_{fg}=\operatorname{Concat}(\sigma_{fg}^{(1)},\ldots,\sigma_{fg}^{(N)}),\\
S_{bg}=\operatorname{Concat}(\sigma_{bg}^{(1)},\ldots,\sigma_{bg}^{(N)})
\end{aligned}
\label{eq:body-20}
\end{equation}

The weight of each prototype is generated from its mean dispersion score:
\begin{equation}
\omega_{fg}^{(i)}=\frac{1}{1+\operatorname{mean}(S_{fg}^{(i)})},
\omega_{bg}^{(i)}=\frac{1}{1+\operatorname{mean}(S_{bg}^{(i)})}\label{eq:body-21}
\end{equation}

After $L_2$ normalization, the foreground-background similarity is computed as
\begin{equation}
A_{cross}=\operatorname{Norm}(V_{fg})\operatorname{Norm}(V_{bg})^\top\label{eq:body-22}
\end{equation}

The foreground-background separation term is defined as
\begin{equation}
\mathcal{L}_{neg}=
\operatorname{mean}\left[
(\omega_{fg}\omega_{bg}^{\top})\odot \exp(A_{cross})
\right]\label{eq:body-23}
\end{equation}

For within-group attraction, VACL forms foreground-foreground and background-background prototype pairs within each batch, where groups refer to foreground/background roles in each episode. Pairs with lower similarity receive higher rank-based attention, summarized as
\begin{equation}
\begin{aligned}
\mathcal{L}_{pos}^{fg}
=\operatorname{mean}\left[
\omega_{fg}^{ij}\exp(-\gamma r_{fg}^{ij})(1-A_{fg}^{ij})
\right],
\\
\mathcal{L}_{pos}^{bg}
=\operatorname{mean}\left[
\omega_{bg}^{ij}\exp(-\gamma r_{bg}^{ij})(1-A_{bg}^{ij})
\right]
\end{aligned}
\label{eq:body-24}
\end{equation}

Here, $A_{fg}^{ij}$ and $A_{bg}^{ij}$ are within-group cosine similarities, $r^{ij}$ is the ascending similarity rank, $\gamma>0$ controls rank decay, and $\omega^{ij}$ denotes the pairwise dispersion-based weight. The final VACL loss is
\begin{equation}
\mathcal{L}_{vacl}
=\mathcal{L}_{neg}
+\frac{1}{2}(\mathcal{L}_{pos}^{fg}+\mathcal{L}_{pos}^{bg})\label{eq:body-25}
\end{equation}

If the batch contains no valid positive pairs, VACL returns a zero loss.

\subsection{Loss Computation}

During training, binary cross-entropy with the smoothed query target $\widetilde{M}_q$ is used as the main segmentation loss:
\begin{equation}
\mathcal{L}_{seg}=
\operatorname{BCE}(\hat{M}_q,\widetilde{M}_q)\label{eq:body-26}
\end{equation}

The total loss combines segmentation supervision with the auxiliary VACL constraint:
\begin{equation}
\mathcal{L}_{total}
=\mathcal{L}_{seg}
+\lambda_{vacl}\mathcal{L}_{vacl}\label{eq:body-27}
\end{equation}

Here, $\lambda_{vacl}$ controls the auxiliary contrastive loss weight. $\mathcal{L}_{seg}$ enforces pixel-level consistency, while $\mathcal{L}_{vacl}$ regularizes foreground/background relations according to variance-derived prototype uncertainty. VACL is used during source-domain training and is disabled during inference.

\begin{table*}[tbp]
\centering
\settowidth{\rankcolumnwidth}{5-shot}
\addtolength{\rankcolumnwidth}{2\tabcolsep}
\resizebox{\textwidth}{!}{
\begin{tabular}{l|l|l|cc|cc|cc|cc|cc}
\hline
\multirow{2}{*}{Method} & \multirow{2}{*}{Venue} & \multirow{2}{*}{Backbone} & \multicolumn{2}{c|}{DeepGlobe} & \multicolumn{2}{c|}{ISIC} & \multicolumn{2}{c|}{Chest X-ray} & \multicolumn{2}{c|}{FSS} & \multicolumn{2}{c}{Mean} \\
\cline{4-13}
& & & 1-shot & 5-shot & 1-shot & 5-shot & 1-shot & 5-shot & 1-shot & 5-shot & 1-shot & 5-shot \\
\hline

SSP & ECCV-22 & ResNet-50 & 40.0 & 48.7 & 35.5 & 45.9 & 74.4 & 74.3 & 78.9 & 80.6 & 57.2 & 62.4 \\
PerSAM\cite{PerSAM} & ICLR-24 & ViT-base & 36.1 & 40.7 & 23.3 & 25.3 & 30.0 & 30.1 & 60.9 & 66.5 & 37.6 & 40.6 \\
PATNet & ECCV-22 & ResNet-50 & 37.9 & 43.0 & 41.2 & 53.6 & 66.6 & 70.2 & 78.6 & 81.2 & 56.1 & 62.0 \\
APM & NeurIPS-24 & ResNet-50 & 40.9 & 44.9 & 41.7 & 51.2 & 78.3 & 82.8 & 79.3 & 81.8 & 60.0 & 65.2 \\
ABCDFSS & CVPR-24 & ResNet-50 & 42.6 & 49.0 & 45.7 & 53.3 & 79.8 & 81.4 & 74.6 & 76.2 & 60.7 & 65.0 \\
DRA & CVPR-24 & ResNet-50 & 41.3 & 50.1 & 40.8 & 48.9 & 82.4 & 82.3 & 79.1 & 80.4 & 60.9 & 65.4 \\
APSeg\cite{APSeg} & CVPR-24 & ViT-base & 35.9 & 40.0 & 45.4 & 54.0 & 84.1 & 84.5 & 79.7 & 81.9 & 61.3 & 65.1 \\
LoEC & CVPR-25 & ResNet-50 & 44.1 & 49.7 & 38.2 & 47.0 & 81.0 & 82.7 & 78.5 & 80.6 & 60.5 & 65.0 \\
SDRC\cite{SDRC} & ICML-25 & ViT-base & 43.2 & 46.8 & 46.6 & 55.0 & 82.9 & 84.8 & 80.3 & 82.6 & 63.2 & 67.3 \\

DFN\cite{DFN}  & ICML-25 & ResNet-50  & 45.7 & 48.0 & 36.3 & 51.1 & \rankcell{rankthird}{85.2} & \rankcell{ranksecond}{90.3} & 80.7 & \rankcell{rankfirst}{85.8} & 62.0 & 68.8 \\
DFN  & ICML-25 & ViT-base   & 39.5 & 47.7 & 50.4 & 58.5 & 83.2 & \rankcell{rankthird}{87.1} & \rankcell{rankfirst}{83.0} & \rankcell{ranksecond}{85.7} & 64.0 & 69.8 \\
IFA  & CVPR-24 & ResNet-50  & \rankcell{rankfirst}{50.6} & \rankcell{rankfirst}{58.8} & \rankcell{ranksecond}{66.3} & \rankcell{ranksecond}{69.8} & 74.0 & 74.6 & 80.1 & 82.4 & \rankcell{rankthird}{67.8} & \rankcell{rankthird}{71.4} \\

HSL & AAAI-26 & ResNet-50 & \rankcell{rankthird}{46.1} & \rankcell{rankthird}{53.8} & 48.0 & 55.6 & 84.6 & 85.3 & 78.2 & 80.4 & 64.2 & 68.8 \\
HSL & AAAI-26 & ViT-base & 45.8 & \rankcell{ranksecond}{54.6} & \rankcell{rankthird}{59.4} & \rankcell{rankthird}{64.6} & \rankcell{ranksecond}{86.0} & 86.3 & \rankcell{ranksecond}{81.9} & \rankcell{rankthird}{83.8} & \rankcell{ranksecond}{68.2} & \rankcell{ranksecond}{72.3} \\
\hline
DAUPNet & Ours & ResNet-50 & \rankcell{ranksecond}{46.74} & 49.98 & \rankcell{rankfirst}{74.41} & \rankcell{rankfirst}{83.02} & \rankcell{rankfirst}{91.50} & \rankcell{rankfirst}{91.93} & \rankcell{rankthird}{80.89} & 83.11 & \rankcell{rankfirst}{73.39} & \rankcell{rankfirst}{77.01} \\
\hline
\end{tabular}
}
\begingroup
\setlength{\fboxsep}{1pt}
\caption{Comparison with state-of-the-art methods on four CD-FSS target domains under 1-shot / 5-shot mIoU (\%). Compared methods use their respective adaptation strategies. The top three values in each column are bold and color-coded: \protect\colorbox{rankfirst}{\textbf{1st}}, \protect\colorbox{ranksecond}{\textbf{2nd}}, and \protect\colorbox{rankthird}{\textbf{3rd}}.}
\label{4-5}
\endgroup
\end{table*}
\section{Experiments and Result Analysis}
\subsection{Datasets and Evaluation Metrics}
 DAUPNet is evaluated under the standard CD-FSS protocol. The model is trained on PASCAL VOC 2012\cite{pascal} with SBD extended annotations as the source domain and evaluated on four unseen target domains: DeepGlobe \cite{deepglobe}, ISIC2018 \cite{ISIC}, the Chest X-ray benchmark \cite{PATNet}, and FSS-1000 \cite{fss-1000}. These domains provide comprehensive coverage of imaging modalities, object structures, and foreground-background ratios under substantial domain shifts.

Evaluation uses both 1-way 1-shot and 1-way 5-shot settings. Each episode contains a support set with annotations and a query image; the model predicts the target region conditioned on the support. Performance is measured by mean intersection over union (mIoU). 

For each evaluation seed, 1,200, 2,400, 1,200, and 2,400 episodes are sampled from DeepGlobe, ISIC2018, Chest X-ray, and FSS-1000, respectively. Inputs are resized to $416\times416$ for ISIC and $400\times400$ for the other domains. Foreground intersections and unions are accumulated within each class, and class IoUs are averaged equally. Reported results average five evaluation seeds (0--4) for a single trained model. The 1-shot and 5-shot evaluations share queries; 1-shot uses the first of the five support images.

\subsection{Implementation Details} DAUPNet uses an ImageNet-pretrained ResNet-50 \cite{ResNet-50} with a frozen encoder. Experiments use two NVIDIA RTX 3090 24GB GPUs, Python 3.9.25, PyTorch 2.5.1, and CUDA 12.1. Training is performed on PASCAL VOC 2012 + SBD \cite{SBD} for 20 epochs using AdamW, batch size 4, learning rate $10^{-3}$, weight decay $5\times10^{-4}$, and a three-epoch warm-up followed by cosine decay. The checkpoint is selected by source-validation foreground IoU. Training uses BCE with 0.02 label smoothing and enables VACL after warm-up. Training augmentation comprises resizing and random horizontal flipping.

The source-trained model is evaluated directly on target domains without parameter updates. Target support masks are used only to construct episode-specific prototypes; target query labels are used only for evaluation. The baseline, ablations, and full model use the same evaluation episodes and metric.

\begin{figure*}[t]
\centering
\includegraphics[width=\linewidth]{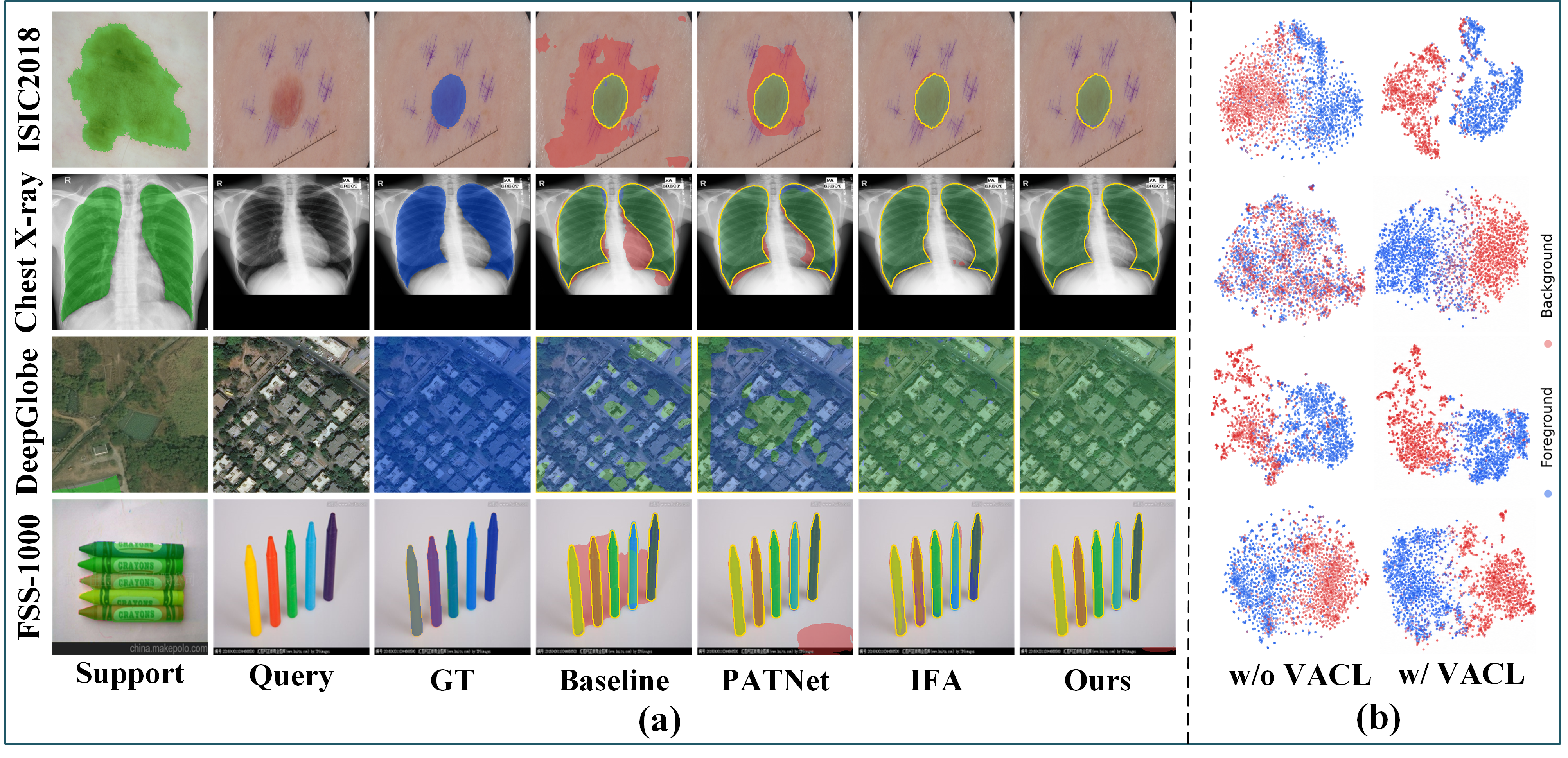} 
\caption{(a) Qualitative comparison of support-guided prediction overlays and GT contours. (b) Foreground/background feature distributions without and with VACL.}
\label{fig:4.1}  
\end{figure*}

\begin{table*}[t]
\centering
\resizebox{\textwidth}{!}{%
\begin{tabular}{clccccccccc}
\toprule
Exp. & Setting & HFH & PPR & VACL & DeepGlobe & ISIC & Chest X-ray & FSS & Mean & Delta \\
\midrule
1 & Baseline &  &  &  & 41.53 & 68.63 & 78.31 & 74.45 & 65.73 & 0.00 \\
2 & +HFH & \checkmark &  &  & 43.42 & 70.86 & 87.95 & 76.57 & 69.70 & +3.97 \\
3 & +HFH+PPR & \checkmark & \checkmark &  & 45.19 & 71.41 & 89.96 & 77.16 & 70.93 & +5.20 \\
4 & +HFH+PPR+VACL & \checkmark & \checkmark & \checkmark & 46.74 & 74.41 & 91.50 & 80.89 & 73.39 & +7.66 \\
\bottomrule
\end{tabular}
}
\caption{Progressive ablation of DAUPNet modules under the 1-shot setting. Delta denotes the mean improvement over Exp.~1.}
\label{4-6}
\end{table*}

\subsection{Comparison with Existing Methods}
\subsubsection{Quantitative Comparison with SOTA Methods}

As shown in Table \ref{4-5}, DAUPNet achieves 73.39\%/77.01\% average mIoU with ResNet-50 in the 1-shot/5-shot settings, yielding average mIoU improvements of 5.59 and 5.61 mIoU points over the strongest prior ResNet-50 results. More importantly, the dataset-wise pattern reveals how DAUPNet uses support evidence under different forms of domain shift. On ISIC, it surpasses the strongest same-backbone results by 8.11 and 13.22 mIoU points under the 1-shot and 5-shot settings, respectively, while its own performance increases from 74.41\% to 83.02\%. This 8.61 mIoU-point improvement suggests that additional support evidence benefits DAUPNet when lesion appearance and boundaries vary substantially. In contrast, Chest X-ray already reaches 91.50\% with one support and improves by only 0.43 mIoU points with five, suggesting that HFH and probabilistic prototypes can recover reliable correspondence from a single example when anatomical structure is regular. The corresponding 1-to-5-shot mIoU improvements on DeepGlobe and FSS-1000 are 3.24 and 2.22 mIoU points, respectively, indicating that additional annotations alone do not equally resolve texture- and shape-dominated variation. Taken together, these results support the proposed reliability chain: DAUPNet converts available support evidence into variance-aware prototype statistics, rather than deriving its gains merely from backbone capacity or a larger support set.

\subsubsection{Qualitative Prediction Comparison}

Fig.\ref{fig:4.1} (a) shows that DAUPNet produces more compact, boundary-consistent masks than the baseline and competing methods. It suppresses false positives in cluttered backgrounds, preserves bilateral structures in Chest X-ray, and follows weak lesion contours more closely in ISIC. Rather than attributing each visual pattern to an isolated module, these results illustrate the complete reliability chain: harmonized features provide coherent spatial evidence, probabilistic prototypes retain ambiguous foreground-background structure, and uncertainty-aware optimization yields a cleaner decision margin. The visual behavior is consistent with the medical-domain advantages over prior same-backbone methods in Table \ref{4-5}. The prediction overlays further show that differences are concentrated around lesion contours and background distractors, where a small representation error can flip the foreground-background decision. This localization of improvements links the visual evidence directly to the medical-domain advantage in Table \ref{4-5}.

\begin{table}[t]
\centering
\setlength{\tabcolsep}{3pt}
\resizebox{\linewidth}{!}{%
\begin{tabular}{ccccccc}
\toprule
 $\lambda_{vacl}$ & DeepGlobe & ISIC & Chest X-ray & FSS & Mean & $\Delta$Mean\\
\midrule
0.000 & 45.19 & 71.41 & 89.96 & 77.16 & 70.93 & 0.00 \\
0.005 & 45.74 & 72.41 & 89.89 & 78.73 & 71.69 & +0.76 \\
0.010 & 45.79 & 72.85 & 90.94 & \textbf{81.04} & 72.66 & +1.73 \\
0.020 & \textbf{46.74} & \textbf{74.41} & \textbf{91.50} & 80.89 & \textbf{73.39} & \textbf{+2.46} \\
0.050 & 44.79 & 70.90 & 90.99 & 77.39 & 71.02 & +0.09 \\
\bottomrule
\end{tabular}
}
\caption{Sensitivity analysis of the VACL loss weight \(\lambda_{vacl}\).}
\label{tab:4-10}
\end{table}  

\begin{figure}[t]
\centering
\includegraphics[width=\linewidth]{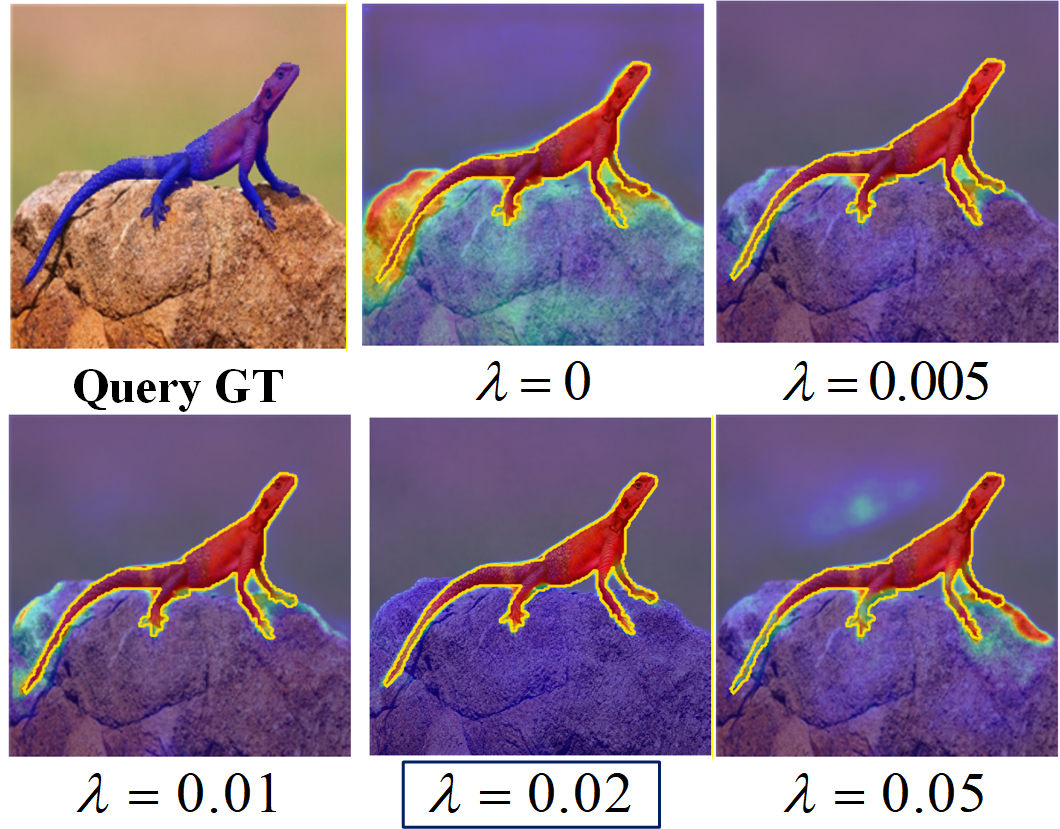} 
\caption{Prediction Responses under Different VACL Weight Coefficients}
\label{lambda_visualization}  
\end{figure}

\subsection{Ablation Study}
   Table \ref{4-6} validates the proposed reliability chain. HFH first raises the mean mIoU from 65.73\% to 69.70\%, showing that stabilizing hierarchical evidence is an effective foundation for prototype estimation. PPR then improves the mean to 70.93\% through multi-level foreground/background prototype construction; its dispersion scores are used only when VACL is enabled. Finally, VACL uses that estimate during optimization and reaches 73.39\%, resulting in an overall improvement of 7.66 mIoU points over the baseline. The domain-wise results are also consistent with this interpretation: HFH improves Chest X-ray by 9.64 mIoU points; PPR further improves ISIC and Chest X-ray by 0.55 and 2.01 mIoU points, respectively; and VACL contributes an additional 3.73 mIoU points on the visually diverse FSS-1000 benchmark. The progressive gains support the combined design, while the supplementary diagnostic shows that VACL also benefits deterministic prototypes.

\subsection{VACL Weight Sensitivity}
Table \ref{tab:4-10} shows that the uncertainty-aware constraint is most effective as a moderate regularizer. Increasing $\lambda _ { v a c l }$ from 0 to 0.020 raises the four-domain mean from 70.93\% to 73.39\%, whereas a weight of 0.050 reduces it to 71.02\%. Excessive contrastive weighting can overemphasize episode-level prototype relations at the expense of pixel-wise supervision. Among the tested weights, FSS-1000 achieves its highest mIoU at $\lambda_{vacl}=0.010$, whereas the other three domains perform best at 0.020, indicating domain-dependent sensitivity to the contrastive weight. The full model uses $\lambda _ { v a c l } = 0.020$.

\subsection{VACL Representation Visualization}
Fig.\ref{fig:4.1} (b) provides a feature-space view of VACL. Compared with HFH+PPR, the full model produces clearer foreground/background separation in target domains, illustrating the feature separation obtained with the combined VACL design. The response maps under different weights further agree with Fig. \ref{lambda_visualization}: $\lambda _ { v a c l } = 0.02$ shows clearer foreground-background discrimination in the displayed examples.

Additional PPR/VACL ablations, domain-distance analysis, and efficiency results are provided in the supplementary appendix (Tables~\ref{4-7}--\ref{tab:ppr-hierarchy}).

 \section{Conclusion}This paper reframes CD-FSS as a problem of maintaining reliable prototype discrimination under domain shift. DAUPNet implements this idea as a coherent chain: HFH stabilizes the hierarchical evidence used to construct prototypes, PPR exposes residual foreground-background uncertainty through probabilistic representation, and VACL weights prototype comparisons using variance-derived dispersion scores. Across four target domains, DAUPNet reaches 73.39\%/77.01\% average mIoU in the 1-shot/5-shot settings and improves the 1-shot baseline by 7.66 mIoU points. Its performance on weak-boundary and grayscale medical images supports the practical value of uncertainty-aware prototype modeling, while the smaller gain on DeepGlobe suggests that texture and appearance variation remain challenging. These findings establish prototype reliability as a useful organizing principle for robust CD-FSS and motivate future work on jointly adapting uncertainty estimation and global domain alignment.

{\small
\bibliographystyle{daupnet_refs}
\bibliography{myref}
}

\clearpage
\appendix
\setcounter{page}{1}
\maketitlesupplementary
\setcounter{figure}{0}
\setcounter{table}{0}
\setcounter{equation}{0}
\renewcommand{\thefigure}{S\arabic{figure}}
\renewcommand{\thetable}{S\arabic{table}}
\renewcommand{\theequation}{S\arabic{equation}}
\section{Appendix}
\begin{figure}[t]
\centering
\includegraphics[width=\linewidth]{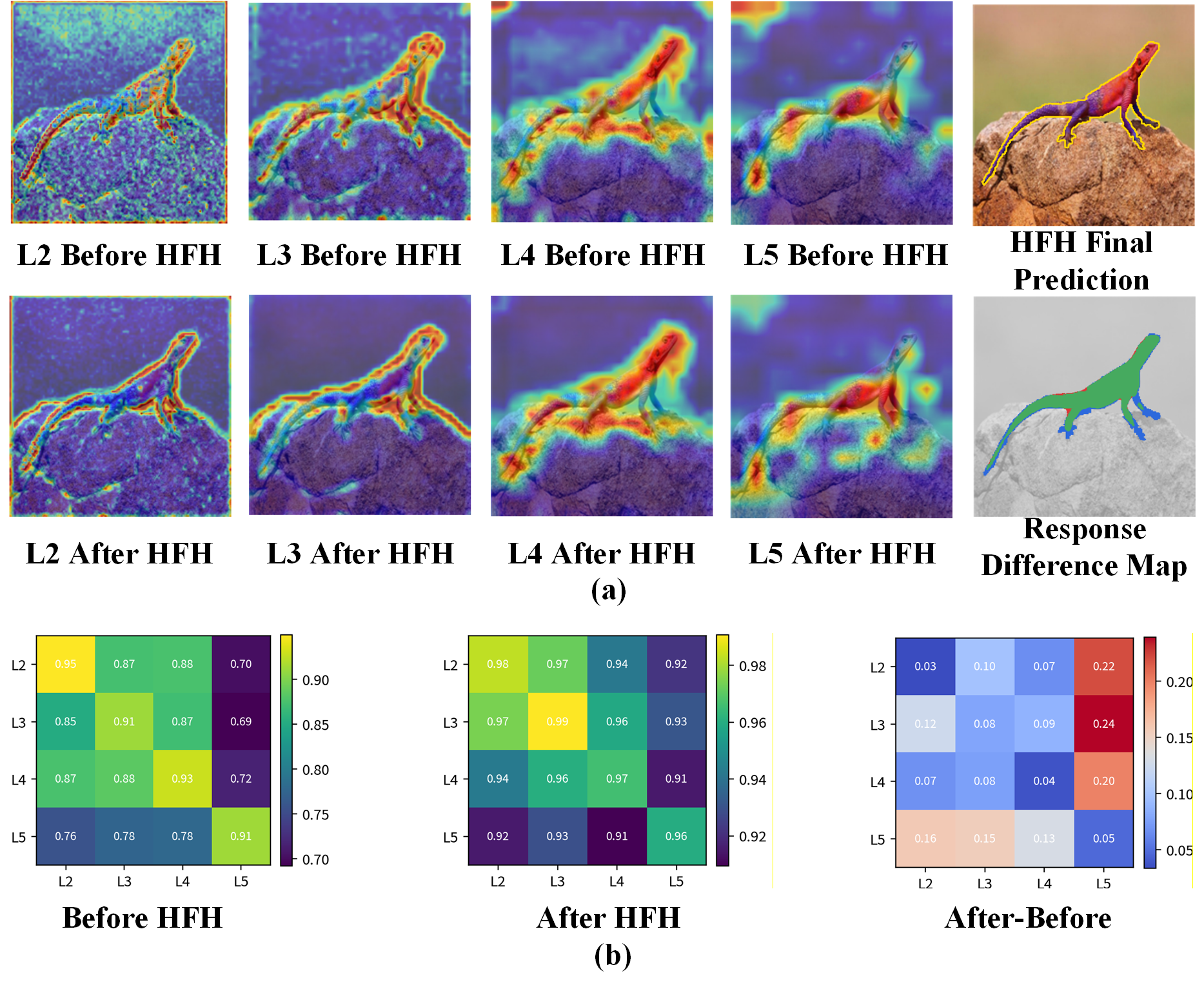} 
\caption{(a) Hierarchical response comparison before and after HFH; (b) Support-query hierarchical cross-similarity matrix before and after HFH.}
\label{4.2}  
\end{figure}
\subsection{HFH Mechanism Visualization}
To further examine how Hierarchical Feature Harmonization (HFH) reorganizes multi-level features before prototype computation, the hierarchical responses and support-query cross-similarity of a representative FSS-1000 episode are visualized.

As shown in Fig.\ref{4.2}(a), the query-branch responses at levels L2--L5 before and after HFH are compared, together with the final prediction and response difference map. Before HFH, responses across different levels are not fully aligned: shallow or mid-level features may retain local texture activations, while deep-level responses are semantically stronger but spatially coarser. After HFH, activations concentrate more on the target region and are less affected by scattered background textures, illustrating the spatial response changes introduced by hierarchical coordination in this example.

As shown in Fig.\ref{4.2}(b), the support-query hierarchical cross-similarity matrix for the same episode is further visualized. Each cell represents the cosine similarity between one support\(-\)layer feature and one query\(-\)layer feature, with rows denoting support layers and columns denoting query layers. The Before HFH and After HFH matrices show the support-query correspondence before and after hierarchical coordination, while the After \(-\) Before difference matrix reveals per\(-\)cell changes introduced by HFH. Positive values in the difference matrix indicate enhanced matching between corresponding support/query layers, while negative values indicate suppressed or weakened matching. In this example, HFH increases overall cross-level support-query similarity. These results are consistent with Table \ref{4-6} in the main text, where HFH raises the mean mIoU from 65.73 to 69.70, and provide the feature-level basis for subsequent PPR prototype statistics.

\subsection{PPR Prototype Modeling Ablation}
To determine which prototype statistics best describe support samples, a controlled ablation of PPR prototype modeling strategies is conducted on FSS-1000, since this domain exhibits substantial variation in category morphology and foreground-background relationships, making it more sensitive to prototype quality.

\begin{table}[t]
\centering
\setlength{\tabcolsep}{10pt}  
\resizebox{\linewidth}{!}{%
\begin{tabular}{cccc}
\toprule
 Exp. & Prototype Source & Variance & mIoU \\
\midrule
1 & Single deterministic &  & 77.30 \\
2 & Deepest-layer fg/bg &  & 78.03 \\
3 & Deepest-layer fg/bg & \checkmark & 79.32 \\
4 & Multi-level fg/bg &  & 78.18 \\
5 & Multi-level fg/bg & \checkmark & \textbf{80.89} \\
\bottomrule
\end{tabular}
}
\caption{Ablation results of PPR prototype modeling strategies.}
\label{4-7}
\end{table}

\begin{figure}[t]
\centering
\includegraphics[width=\linewidth]{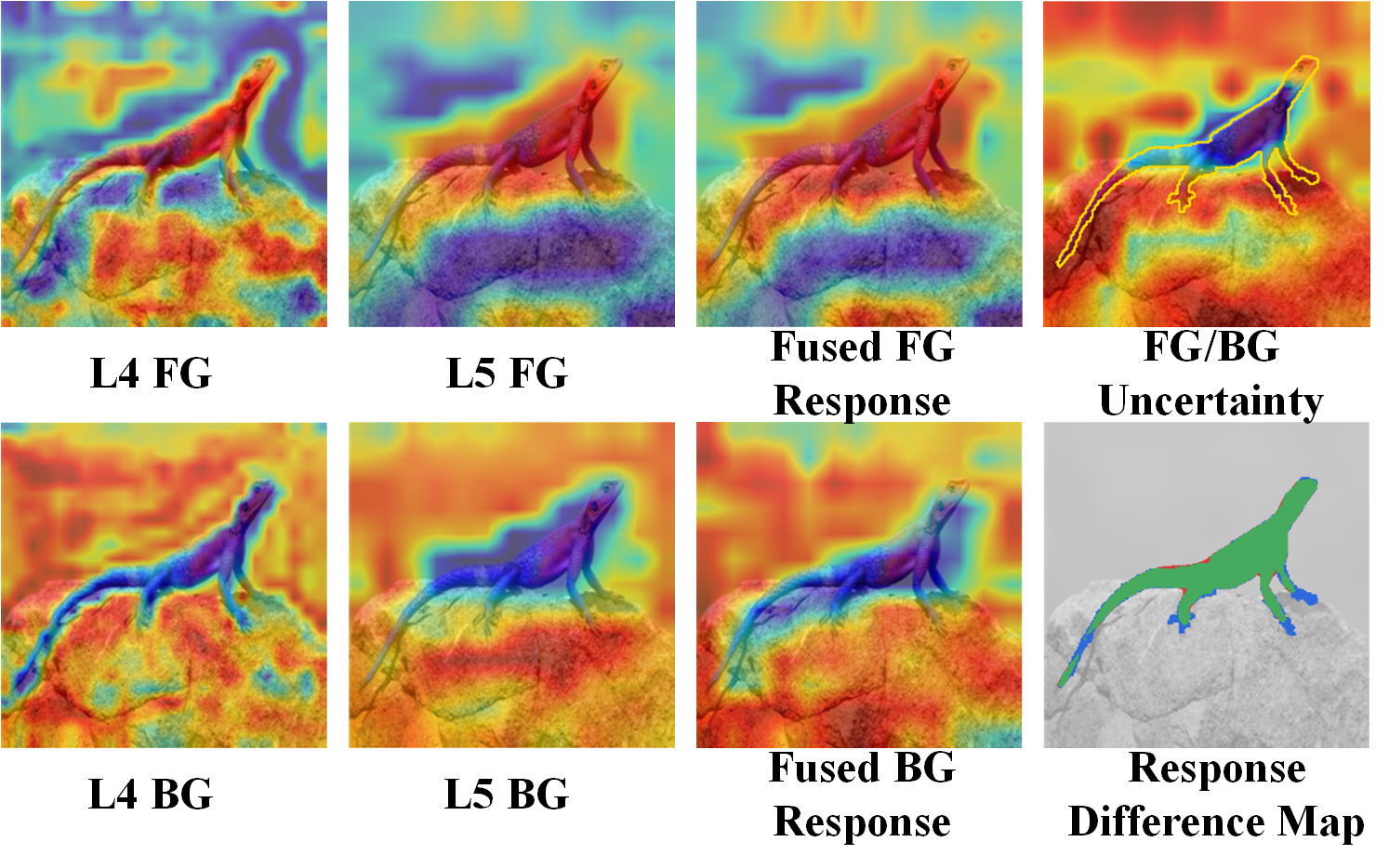} 
\caption{Hierarchical foreground/background prototype response reorganization of PPR.}
\label{4.3}  
\end{figure}

\begin{table*}[t]
\centering
\setlength{\tabcolsep}{4pt}
\resizebox{\linewidth}{!}{%
\begin{tabular}{ccccc}
\toprule
Exp. & \multicolumn{1}{c}{Setting} & Representation Path & \multicolumn{1}{c}{VACL Input}& Four-Domain Mean \\
\midrule
1 & Baseline & baseline feature & no VACL & 65.73 \\
2 & Baseline + VACL & baseline feature & deterministic prototypes & 66.67 \\
3 & HFH + VACL w/o PPR & HFH feature & HFH deterministic prototypes & 70.21 \\
4 & HFH + PPR + VACL & HFH + PPR feature & PPR probabilistic prototypes and variances & \textbf{73.39} \\
\bottomrule
\end{tabular}
}
\caption{Diagnostic results on the dependency between PPR and VACL.}
\label{4-8}
\end{table*}

\begin{table*}[t]
\centering
\setlength{\tabcolsep}{4pt}
\resizebox{\linewidth}{!}{%
\begin{tabular}{@{}clcccccc@{}}
\toprule
Exp. & Setting & Variance Weight & Rank Weight & FG-BG Separation & FG Attraction & BG Attraction & mIoU \\
\midrule
1 & w/o VACL &  &  &  &  &  & 77.16 \\
2 & Standard contrastive &  &  & \checkmark & \checkmark & \checkmark & 78.42 \\
3 & Variance-aware contrastive & \checkmark & \checkmark & \checkmark & \checkmark & \checkmark & \textbf{80.89} \\
4 & Negative only & \checkmark & \checkmark & \checkmark &  &  & 78.72 \\
5 & Positive only & \checkmark & \checkmark &  & \checkmark & \checkmark & 79.59 \\
6 & FG positive + negative & \checkmark & \checkmark & \checkmark & \checkmark &  & 79.60 \\
\bottomrule
\end{tabular}
}
\caption{Ablation results of VACL components and loss terms.}
\label{4-9}
\end{table*}

\begin{figure*}[t]
\centering
\includegraphics[width=\linewidth]{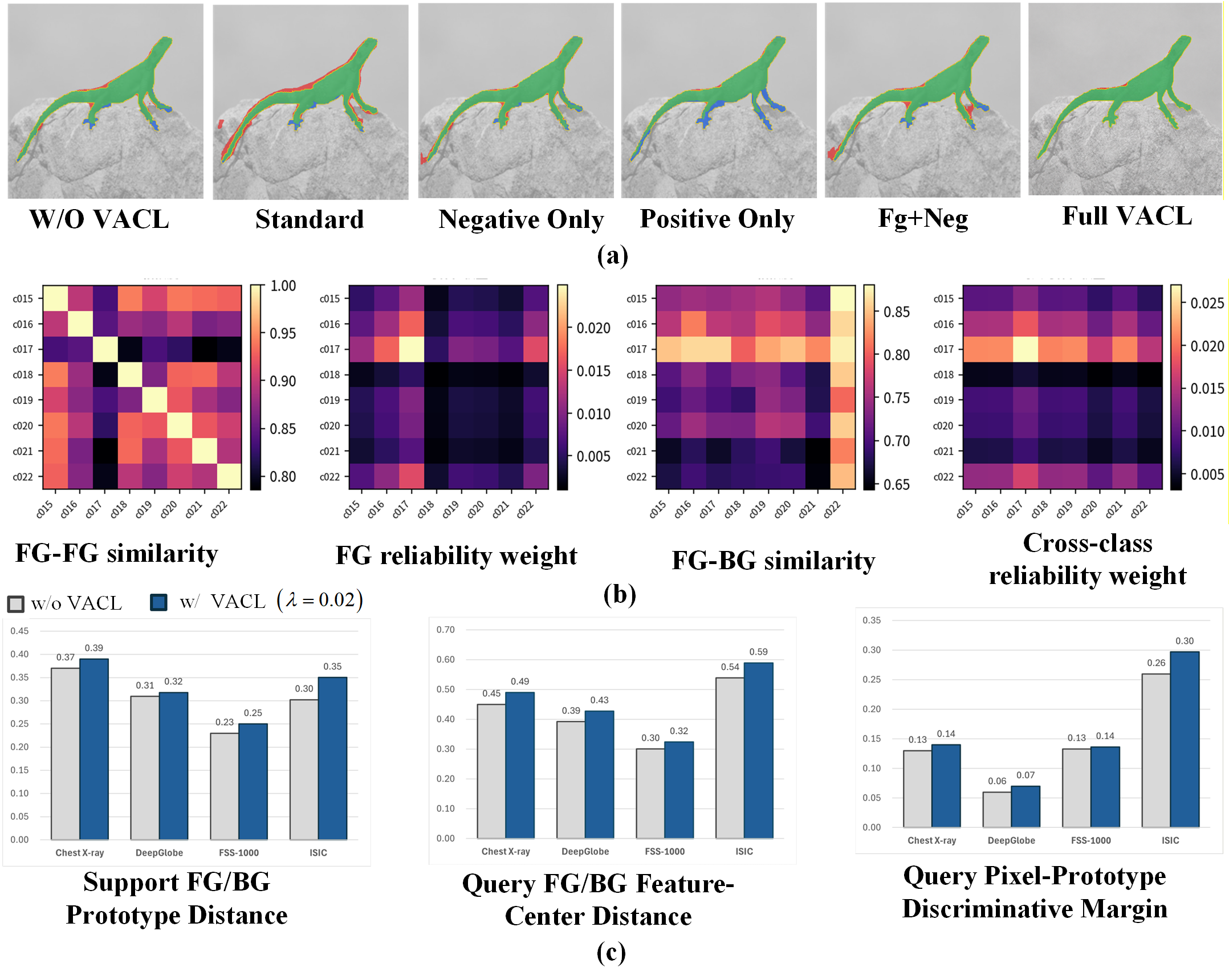} 
\caption{(a) Prediction comparison for VACL component ablation; (b)  VACL reliability relation matrices (FG-FG similarity, FG reliability weight, FG-BG similarity, cross-class reliability weight); (c) Foreground/background separation metrics (support prototype distance, query feature-center distance, query pixel-prototype margin).}
\label{4.4}  
\end{figure*}

As shown in Table \ref{4-7}, five configurations with distinct prototype sources and variance estimation strategies are compared: simple deterministic prototype (Exp. 1), deepest-layer foreground/background prototypes without variance (Exp. 2), deepest-layer foreground/background prototypes with variance (Exp. 3), multi-level foreground/background prototypes without variance (Exp. 4), and full PPR with multi-level prototypes and variance estimation (Exp. 5). This comparison is organized around two factors: prototype source granularity (single vs. deepest-layer vs. multi-level) and whether variance estimation is incorporated. Since variance estimation depends on foreground/background prototype statistics, not all column combinations are semantically meaningful. HFH and VACL are retained in these controls; the single deterministic configuration in Exp.~1 is therefore distinct from the baseline in Table \ref{4-6}.

As shown in Table \ref{4-7}, the benefit of PPR primarily stems from changes in prototype statistics. Exp. 5 achieves the best result in this group (80.89), exceeding the deepest-layer variant with variance (Exp. 3, 79.32) and the multi-level variant without variance (Exp. 4, 78.18). Exp. 4 and Exp. 5 further indicate that incorporating variance estimation improves the multi-level configuration by 2.71 mIoU points under the same HFH and VACL setting. These results support combining multi-level foreground/background statistics with variance-aware weighting.

As shown in Fig.\ref{4.3}, the PPR prototype modeling process is decomposed into interpretable intermediate states on the same FSS-1000 episode. The figure presents L4/L5 foreground and background responses, PPR-fused foreground and background responses, a foreground/background response ambiguity map, and a response difference map. L4 responses better preserve edges, local textures, and fine-grained shape information, while L5 responses emphasize semantic targets but are spatially coarser. Consequently, single-level prototypes struggle to simultaneously capture target localization and boundary recovery. After PPR reorganizes multi-level foreground/background statistics into probabilistic prototypes, the fused foreground response indicates whether target regions are stably activated, and the fused background response reveals whether non-target regions and distracting textures are effectively suppressed. The panel labeled FG/BG Uncertainty is constructed from the normalized difference between fused foreground and background responses: similar responses indicate greater response ambiguity. This map visualizes response competition rather than the learned prototype-level dispersion scores. The final response difference map uses predictions from the HFH+PPR checkpoint (without VACL) to show residual false positives and false negatives. This visualization illustrates how paired foreground/background prototypes reorganize responses and where foreground and background responses compete in this episode.

\subsection{PPR-VACL Dependency Diagnostic}
To verify whether the probabilistic prototypes and variance estimates provided by PPR effectively support VACL, a diagnostic experiment is conducted with two types of controls: first, adding VACL without full PPR to test whether the contrastive constraint has independent effects; second, comparing deterministic prototypes against PPR probabilistic prototypes to examine their combined effect with VACL.

As shown in Table \ref{4-8}, the configurations are organized by the input source to VACL, focusing on how different prototype statistics support contrastive constraints.

As shown in Table \ref{4-8}, adding VACL under deterministic prototype conditions also improves performance, indicating that VACL does not require full PPR to take effect. Baseline+VACL improves over Baseline from 65.73 to 66.67. With HFH fixed, VACL improves the mean from 69.70 (Table~\ref{4-6}) to 70.21, showing that the contrastive constraint can benefit deterministic prototypes. The full HFH+PPR+VACL configuration further introduces probabilistic prototypes and variance estimates from PPR, providing a reliability-aware basis for contrastive weighting and reaching a four-domain mean of 73.39. These results support combining PPR and VACL, which weights prototype relations using dispersion-derived scores.

\subsection{VACL Component Ablation}
To further examine VACL's internal design, joint variance/rank weighting and distinct contrastive terms are compared.

As shown in Table \ref{4-9}, six configurations organized from two perspectives are compared: first, comparing standard contrastive learning against the joint variance/rank-weighted design to assess their combined effect; second, decomposing VACL loss terms into foreground-background separation, foreground within-group attraction, and background within-group attraction to analyze the contributions of different contrastive relations. All experiments are conducted on FSS-1000.

As shown in Table \ref{4-9}, full VACL reaches 80.89 mIoU, improving over w/o VACL (77.16) and standard contrastive learning (78.42) by 3.73 and 2.47 points, respectively. Since the full configuration enables both variance and rank weighting, this comparison supports their combined effect. Negative only, positive only, and FG positive + negative achieve 78.72, 79.59, and 79.60, respectively, all below the full configuration. These results indicate that combining foreground-background separation with both within-group attraction terms is beneficial in this FSS-1000 experiment.

As shown in Fig.\ref{4.4}, the VACL analysis is organized across four dimensions: component ablation, weight response, reliability relations, and feature separation. Fig.\ref{4.4}(a) compares w/o VACL, standard contrastive, negative only, positive only, FG positive + negative, and full VACL on the same episode using high-contrast error maps to observe how pixel-level errors migrate after introducing different loss terms. Green, red, and blue respectively indicate regions where predictions align with, overshoot, or undershoot the ground truth, enabling analysis of VACL's impact on boundaries, background suppression, and foreground coverage. Fig.\ref{4.4}(b) characterizes the weighting basis of VACL through inter-episode prototype similarity and reliability weights, including FG-FG similarity (cosine similarity between foreground prototypes across episodes), FG reliability weight (pairwise combinations of foreground prototype reliability weights), FG-BG similarity (cross-class similarity between foreground and background prototypes), and cross-class reliability weight (combined weighting strength of foreground and background reliability). Fig.\ref{4.4}(c) quantifies changes in the foreground/background representation structure by comparing support prototype distance, query feature-center distance, and query pixel-prototype margin with VACL disabled versus enabled ($\lambda_{\text{vacl}} = 0.02$) across all four target domains.

\subsection{Domain Distance and Target-Domain Performance}
To further analyze the degree of domain shift across different target domains, image features are extracted from the source domain (PASCAL VOC) and four target domains, and RBF-kernel MMD with L2 normalization is adopted as the domain-distance metric. This analysis does not participate in model training; it provides a descriptive comparison of source-target feature distributions under the stated feature extraction and kernel settings.

\begin{table}[t]
\centering
\setlength{\tabcolsep}{4pt}
\setlength{\tabcolsep}{2pt}  
\resizebox{\linewidth}{!}{%
\begin{tabular}{@{}lcccc@{}}
\toprule
Target Domain & MMD Distance & Baseline  & DAUPNet  & $\triangle$mIoU \\
\midrule
DeepGlobe & 0.3821 & 41.53 & \textbf{46.74} & 5.21 \\
ISIC & 0.4546 & 68.63 & \textbf{74.41} & 5.78 \\
Chest X-ray & 0.4906 & 78.31 & \textbf{91.50} & \textbf{13.19} \\
FSS-1000 & 0.2072 & 74.45 & \textbf{80.89} & 6.44 \\
\bottomrule
\end{tabular}
}
\caption{MMD distance from the source domain to target domains and 1-shot performance.}
\label{4-11}
\end{table}

\setlength{\dbltextfloatsep}{8pt plus 2pt minus 2pt}
\setlength{\dblfloatsep}{4pt plus 1pt minus 1pt}
\begin{table*}[!t]
\centering
\renewcommand{\arraystretch}{0.9}
\setlength{\tabcolsep}{4pt}
\begin{tabular}{@{}lccccc@{}}
\toprule
Method & Input Size & Params (M) & FLOPs (G) & FPS & Memory (MiB) \\
\midrule
Baseline & 416 & \textbf{28.100673} & \textbf{96.035} & \textbf{61.88} & \textbf{323.90} \\
DAUPNet & 416 & 28.439873 & 99.978 & 51.56 & 376.13 \\
Baseline & 400 & \textbf{28.100673} & \textbf{89.555} & \textbf{62.87} & \textbf{305.98} \\
DAUPNet & 400 & 28.439873 & 93.204 & 51.81 & 355.96 \\
\bottomrule
\end{tabular}
\caption{Comparison of parameters, computation, and inference speed. Measurements use 1-shot, FP32, and GPU-resident inputs. FPS is computed from median latency; memory denotes peak allocated MiB. FLOPs count one fused multiply-add as two operations.}
\label{4-12}
\end{table*}

\begin{table*}[!t]
\centering
\renewcommand{\arraystretch}{0.9}
\begin{tabular}{clccccc}
\toprule
Exp. & PPR Configuration & DeepGlobe & ISIC & Chest X-ray & FSS-1000 & Mean \\
\midrule
1 & Deepest + variance & 45.78 & 72.84 & 89.97 & 79.32 & 71.98 \\
2 & Multi-level + variance & \textbf{46.74} & \textbf{74.41} & \textbf{91.50} & \textbf{80.89} & \textbf{73.39} \\
\bottomrule
\end{tabular}
\caption{Deepest-level versus multi-level PPR across four target domains under the 1-shot setting.}
\label{tab:ppr-hierarchy}
\end{table*}

Specifically, given source-domain feature set $X = \{x_i\}$ and target-domain feature set $Y = \{y_j\}$, the squared MMD is computed as
\begin{equation}
\begin{split}
\operatorname{MMD}^2(X,Y) = \frac{1}{n^2} \sum_{i,i'} k(x_i, x_{i'}) &+ \frac{1}{m^2} \sum_{j,j'} k(y_j, y_{j'}) \\- \frac{2}{nm} \sum_{i,j} k(x_i, y_j),
\end{split}
\end{equation}
where $k(a,b) = \exp\left(-\|a-b\|_2^2 / 2\sigma^2\right)$ is the RBF kernel. In the implementation, features are extracted by global average pooling of the deepest feature level of the source-trained model at an input size of 400 pixels and are L2-normalized; up to 1,000 images are used per domain, and the median pairwise Euclidean distance within each pooled source-target sample set is used as the RBF bandwidth $\sigma$; the final $MMD$ distance reported in Table \ref{4-11} is the square root of the non-negative $MMD^2$. $MMD$ values depend on four factors: the feature extractor and selected feature layer, the sample size and category composition of source and target sets, feature normalization, and the RBF kernel bandwidth. Therefore, the $MMD$ values in Table \ref{4-11} provide a descriptive comparison under the same extractor, normalization, and kernel construction rule, with bandwidth estimated separately for each source-target pair, and should not be directly compared with $MMD$ values computed under different configurations in other works.

As shown in Table \ref{4-11}, DeepGlobe exhibits an $MMD$ distance of 0.3821 from the source domain, and its segmentation performance is substantially lower than other target domains, indicating that aerial imaging viewpoints, extensive background textures, and elongated road structures collectively create strong domain shift. ISIC has a larger $MMD$ distance of 0.4546, yet lesion boundaries and low-contrast regions still impose domain-specific medical challenges. FSS-1000 has the smallest $MMD$ distance (0.2072), where DAUPNet achieves a notable gain over the baseline, suggesting that the full model compensates for morphological variations in natural long-tailed categories. Chest X-ray has the largest $MMD$ distance (0.4906), but its target structures are relatively stable and it benefits substantially from the complete modeling chain, yielding the highest mIoU among the four target domains. These descriptive results do not imply that domain distance and performance gains are monotonically related.

\subsection{Parameter Count, Computation, and Inference Efficiency}
Beyond segmentation accuracy, the parameter count, computational cost, inference speed, and memory usage of DAUPNet relative to the baseline are further quantified. This analysis evaluates whether the HFH, PPR, and VACL structures introduce unacceptable computational overhead.

As shown in Table \ref{4-12}, compared to the baseline, DAUPNet introduces only approximately 0.3392M additional parameters, corresponding to a roughly 1.21\% parameter increase. Under input sizes of 416 and 400, DAUPNet adds approximately 3.943G and 3.649G FLOPs, with memory usage increasing by approximately 52.23 MiB and 49.99 MiB, respectively. Inference speed decreases, indicating that hierarchical fusion and prototype inference introduce additional runtime overhead. Nevertheless, in terms of parameter count and memory usage, the complete model remains relatively lightweight.

\subsection{PPR Hierarchy Across Domains}
To further examine whether the multi-level prototype design generalizes beyond FSS-1000, deepest-level and multi-level PPR are compared across all four target domains. HFH and VACL are fixed, and both configurations retain variance estimation under the 1-shot setting.

As shown in Table \ref{tab:ppr-hierarchy}, multi-level PPR reaches a four-domain mean of 73.39, compared with 71.98 for the deepest-level configuration. The improvements on DeepGlobe, ISIC, Chest X-ray, and FSS-1000 are 0.96, 1.57, 1.53, and 1.57 mIoU points, respectively. These results are consistent with Table \ref{4-7} and support combining hierarchical foreground/background statistics under the fixed HFH and VACL setting.

\subsection{Limitations Analysis}
First, the 5-shot results on DeepGlobe fall below some recent methods, indicating that in remote sensing imagery, large-scale repetitive textures and complex backgrounds can still induce foreground-background confusion. This typically manifests as background confusion and boundary expansion, where the model misclassifies background regions with textures similar to the target as foreground.

Second, small objects, elongated structures, and partial occlusions in FSS-1000 increase the difficulty of prototype estimation. When the target region in the support sample is small, or when query-image targets exhibit substantial scale variation, the model may produce missed detections for small objects or under-segmentation of structural regions. Although multi-level prototypes improve average FSS-1000 performance in Tables \ref{4-7} and \ref{tab:ppr-hierarchy}, these aggregate gains do not eliminate the small-object and structural errors observed in individual episodes.

Third, boundary ambiguity and low-contrast regions remain challenging in medical target domains such as ISIC and Chest X-ray. Although DAUPNet achieves high mIoU on medical domains, qualitative results indicate that the model may still produce boundary over-segmentation or local under-segmentation in regions with incomplete lesion edges or grayscale structural boundaries, suggesting room for improvement in boundary calibration.

These observations indicate that the primary failure modes of DAUPNet are concentrated in four categories: small-object missed detection, boundary over-segmentation, background confusion, and structural under-segmentation. Background confusion is associated with the complex remote sensing textures of DeepGlobe; small-object missed detection and structural under-segmentation relate to target scale and morphological variations in FSS-1000; and boundary over-segmentation is also observed in weak-boundary medical images. These observations are consistent with the quantitative results and qualitative prediction figures presented in the main text.

\end{document}